\pdfoutput=1

\documentclass[11pt]{article}

\usepackage{acl}

\usepackage{times}
\usepackage{latexsym}

\usepackage[T1]{fontenc}

\usepackage[utf8]{inputenc}

\usepackage{microtype}

\usepackage{inconsolata}
\usepackage{todonotes}
\usepackage{rotating}
\usepackage{graphicx}
\usepackage{balance} 
\usepackage{lscape}
\usepackage{amsmath}
\usepackage{amssymb}
\usepackage{amsthm}
\usepackage{float}
\usepackage{multirow}
\usepackage{algorithm}
\usepackage{algpseudocode}
\usepackage{tcolorbox}
\usepackage{booktabs}
\usepackage{pifont}
\usepackage{bbm}
\usepackage[normalem]{ulem}
\usepackage{hyperref}
\usepackage{algorithmicx}
\useunder{\uline}{\ul}{}

\algdef{SE}[DOWHILE]{Do}{doWhile}{\algorithmicdo}[1]{\algorithmicwhile\ #1}%

\newtheorem{example}{Example}

\algnewcommand{\LineComment}[1]{\State \(\triangleright\) #1}
\title{SCAIR: Schema-Conditioned Agentic Iterative Reasoning for\\ Enterprise Knowledge Graphs }

\author{%
Prateek Chaturvedi\textsuperscript{1,2},
Yuqicheng Zhu\textsuperscript{1,2}, 
Hongkuan Zhou\textsuperscript{1,2},
Dongzhuoran Zhou\textsuperscript{2,3},\\
\textbf{Yunjie He\textsuperscript{1,2},}
\textbf{Steffen Staab\textsuperscript{1,4},} 
\textbf{Fei Du\textsuperscript{5,6},} 
\textbf{Jie Tang\textsuperscript{6},}
\textbf{Evgeny Kharlamov\textsuperscript{2,3}} 
\\
\textsuperscript{1}University of Stuttgart, 
\textsuperscript{2}Bosch Center for AI,
\textsuperscript{3}University of Oslo,\\
\textsuperscript{4}University of Southampton,
\textsuperscript{5}BSH Home Applications Holding (China) Co., Ltd,\\
\textsuperscript{6}Tsinghua University
\\
\texttt{yuqicheng.zhu@de.bosch.com}
}

\begin{document}
\maketitle
\begin{abstract}
Knowledge Graph–based Retrieval-Augmented Generation (KG-RAG) enables natural language interaction with structured enterprise knowledge, yet existing agentic approaches that perform well on public benchmarks often fail to generalize to real-world enterprise Knowledge Graphs (KGs), which are dense, schema-driven, and operationally constrained. 
To address these limitations, we propose SCAIR (Schema-Conditioned Agentic Iterative Reasoning), a training-free framework that integrates structured planning with controlled iterative reasoning by injecting schema-conditioned structural priors and enforcing schema-aware traversal during multi-hop reasoning. Experiments on an enterprise-oriented benchmark constructed from a real-world Configuration Management DataBase (CMDB) demonstrate that SCAIR substantially improves performance over existing KG-RAG methods.
Crucially, our study highlights that reliable enterprise graph reasoning cannot rely on generic agentic designs; instead, it must explicitly incorporate the target domain's structural and operational constraints into the reasoning process. We demonstrate that by aligning agent design with business logic, substantial performance gains can be achieved without the need for costly model retraining.

\end{abstract}

\section{Introduction}
In the era of Industry 4.0, enterprises generate vast amounts of heterogeneous data distributed across isolated systems, ranging from structured sensor logs to unstructured maintenance reports~\cite{lasi2014industry, frank2019industry}. Knowledge Graphs (KGs), as a practical solution for integrating such data, provide a flexible structure that explicitly models entities and their dependencies within a unified semantic network~\cite{hogan2021knowledge, pan2024unifying}. This structural representation is particularly valuable for complex industrial applications~\cite{DBLP:conf/d2r2/0002HHBIWWBBT23, listl2024knowledge, grangel2020knowledge}. For example, in enterprise IT and manufacturing environments, Configuration Management Databases (CMDBs) are usually modeled as KGs to represent machines, components, production lines, and their operational states~\cite{schmidt2025myam}. Such KGs support business-critical queries, including dependency analysis, fault diagnosis, and component replacement, where reasoning over interconnected assets is essential for reliable decision-making.

To democratize access to structured enterprise knowledge, KG–based Retrieval-Augmented Generation (KG-RAG) has emerged as an effective solution~\cite{peng2024graph, zhu2025knowledge, zhang2025survey}. By enabling users to interact with enterprise KGs through natural language, KG-RAG retrieves grounded and explainable evidence to answer complex queries. More recently, agentic KG-RAG approaches typically follow one of two paradigms: \emph{plan-and-execute}, which learns explicit reasoning plans or subgraph selection strategies \cite{he2024g, DBLP:conf/iclr/LuoLHP24}, and \emph{ReAct-style iterative exploration}, which dynamically interleaves reasoning and graph traversal \cite{sun2023think, chen2024plan, zhou2025gr}. Both paradigms have demonstrated strong performance on public KGQA benchmarks.

To evaluate the deployment readiness of these paradigms, we construct an enterprise-oriented KGQA benchmark derived from a real-world manufacturing CMDB. Unlike the sparse and open-domain graphs used in existing benchmarks, this dataset captures the dense connectivity, strict schema constraints, and operational dependencies characteristic of industrial environments. Experiments on this benchmark reveal that existing agentic KG-RAG methods fail to generalize under these conditions. Public benchmarks can partially obscure these weaknesses, both because their graphs are structurally simpler and because entity and relation names often overlap with LLM pretraining corpora, which may inflate reported performance. In contrast, the enterprise setting exposes systematic failure modes: ReAct-style exploration suffers from uncontrolled search expansion and semantic drift in dense subgraphs, while plan-and-execute methods rely heavily on distribution-specific training and struggle to generalize across query patterns.

Motivated by these findings, we introduce \textbf{SCAIR} (\textbf{S}chema-\textbf{C}onditioned \textbf{A}gentic \textbf{I}terative \textbf{R}easoning), a hybrid agentic framework that unifies structured planning with controlled iterative reasoning. SCAIR injects schema-conditioned structural priors, enforces schema-aware traversal during multi-hop reasoning, and controls topic entity propagation to balance exploration and exploitation during search. 
Notably, SCAIR outperforms all evaluated baselines on the enterprise benchmark without any task-specific training. 

The key lesson from this study is that generic agentic designs optimized on simplified public benchmarks often fail to generalize to real business use cases. Effective enterprise graph reasoning must explicitly account for the structural and operational characteristics of the target domain and reflect these constraints in the agent design. With a clear understanding of the business scenario and systematic failure analysis of existing solutions, substantial performance gains can be achieved without costly model retraining.

\section{Preliminaries}
\subsection{Knowledge Graph Question Answering}
KGs represent structured knowledge as a graph of entities connected by typed relations. Formally, a KG is defined as $\mathcal{G} = (\mathcal{E}, \mathcal{R}, \mathcal{T})$, where $\mathcal{E}$ denotes the set of entities, $\mathcal{R}$ the set of relation types, and $\mathcal{T} \subseteq \mathcal{E} \times \mathcal{R} \times \mathcal{E}$ the set of triples. Each triple $(e_h, r, e_t) \in \mathcal{T}$ encodes an atomic fact, stating that a relation $r \in \mathcal{R}$ holds between a head entity $e_h \in \mathcal{E}$ and a tail entity $e_t \in \mathcal{E}$.
Knowledge Graph Question Answering (KGQA) aims to answer natural language questions by reasoning over relevant facts in $\mathcal{G}$.


\subsection{Existing Approaches}
\label{sec:existing_methods}
KGQA has been addressed through two main paradigms. \textbf{Semantic parsing} approaches translate natural language questions into executable logical forms, such as SPARQL queries, which can then be executed over the KG to obtain answers \cite{DBLP:conf/acl/BerantL14, DBLP:conf/emnlp/BerantCFL13, DBLP:journals/tacl/ReddyLS14}. In contrast, \textbf{information retrieval–based} methods (i.e., KG-RAG) retrieve and rank candidate entities or subgraphs using distributed representations, often leveraging graph neural networks to encode structural information, and then generate answers conditioned on the retrieved evidence \cite{DBLP:journals/corr/BordesUCW15, DBLP:conf/acl/DongWZX15}. More recently, the field has converged toward \textbf{agentic reasoning frameworks}, in which an LLM operates as a decision-making agent that plans, interacts with the KG, and reasons over retrieved evidence \cite{DBLP:conf/iclr/LuoLHP24, he2024g, sun2023think, chen2024plan}.

Within this agentic paradigm, approaches such as Reasoning on Graphs (RoG) \cite{DBLP:conf/iclr/LuoLHP24} and G-Retriever \cite{he2024g} follow a \textbf{Plan-and-Execute} design \cite{DBLP:conf/acl/WangXLHLLL23}. The agent first constructs an explicit reasoning plan or identifies a target subgraph, and subsequently executes this plan through constrained graph retrieval before generating an answer. These methods are primarily training-based, as their planning or retrieval components are learned from supervision to align reasoning plans or subgraph selection with downstream question-answering objectives. Think-on-Graph (ToG) \cite{sun2023think} and Plan-on-Graph (PoG) \cite{chen2024plan} adhere more closely to the \textbf{ReAct}-style framework \cite{DBLP:conf/iclr/YaoZYDSN023}, interleaving reasoning steps with iterative graph exploration and allowing the agent to adapt traversal decisions based on intermediate observations. This class of methods is typically training-free, relying on the LLM’s reasoning capabilities rather than task-specific parameter optimization.
Beyond text-centric KG-RAG, recent works also explore combining knowledge graphs with multimodal signals to improve representation learning and reasoning~\cite{DBLP:conf/iclr/YuTXCRYLWHL025, DBLP:conf/aaai/ZhouHMSZWNS26, DBLP:journals/corr/abs-2410-15981}.
\section{An Enterprise-Oriented KGQA Benchmark}
In this section, we first examine why widely used KGQA benchmarks fail to capture the requirements of industrial applications. We then introduce an enterprise-oriented benchmark constructed from a real-world Configuration Management Database (CMDB), designed to evaluate agentic KG-RAG systems under realistic operational constraints. 

\subsection{Limitations of Existing KGQA}
KGQA benchmarks such as \emph{WebQSP} \cite{Yih2016webqsp} and \emph{Complex Web Questions (CWQ)} \cite{Talmor2018WSQ} are built on open-domain KGs (e.g., Freebase \cite{DBLP:conf/sigmod/BollackerEPST08}) and are designed to evaluate compositional reasoning over relatively sparse graph structures. While influential, these benchmarks rely on assumptions that do not align with enterprise use cases.

First, they assume that each question maps to a single, well-defined executable query that fully captures the user’s intent. In industrial settings, however, questions are often underspecified and constraint-driven, and multiple reasoning paths may be valid depending on the operational context. Second, benchmark questions are largely \emph{retrospective}, treating the KG as an encyclopedia for fact retrieval. Enterprise questions are typically \emph{prospective} and problem-oriented, involving implicit business logic such as operational status, compatibility, or conditional replacement, which is encoded structurally in the KG rather than explicitly stated in text.
Finally, semantic leakage from LLM pretraining is difficult to avoid in public benchmarks, as entity and relation names often overlap with pretraining data. This makes it unclear whether strong performance reflects genuine graph reasoning or implicit memorization. 

These differences expose a clear gap between existing KGQA benchmarks and the requirements of real-world industrial applications.

\subsection{Enterprise KGQA Benchmark}
To address the gap between public benchmarks and real industrial requirements, we construct an enterprise KGQA benchmark grounded in a real-world manufacturing CMDB. The benchmark consists of (i) a KG derived from the CMDB (CMDB-KG), and (ii) business-oriented question–answer pairs.

\paragraph{CMDB-KG.}
CMDB-KG is constructed from a real-world manufacturing CMDB~\cite{DBLP:conf/esws/SchmidtGHWKP25} that integrates heterogeneous enterprise data, including production lines, machines, components, manufacturers, and operational attributes. The graph contains 116,369 triples. Construction details are provided in Appendix~\ref{app:kg_construciton}.

\begin{figure}[t]
    \centering
    \includegraphics[width=\columnwidth]{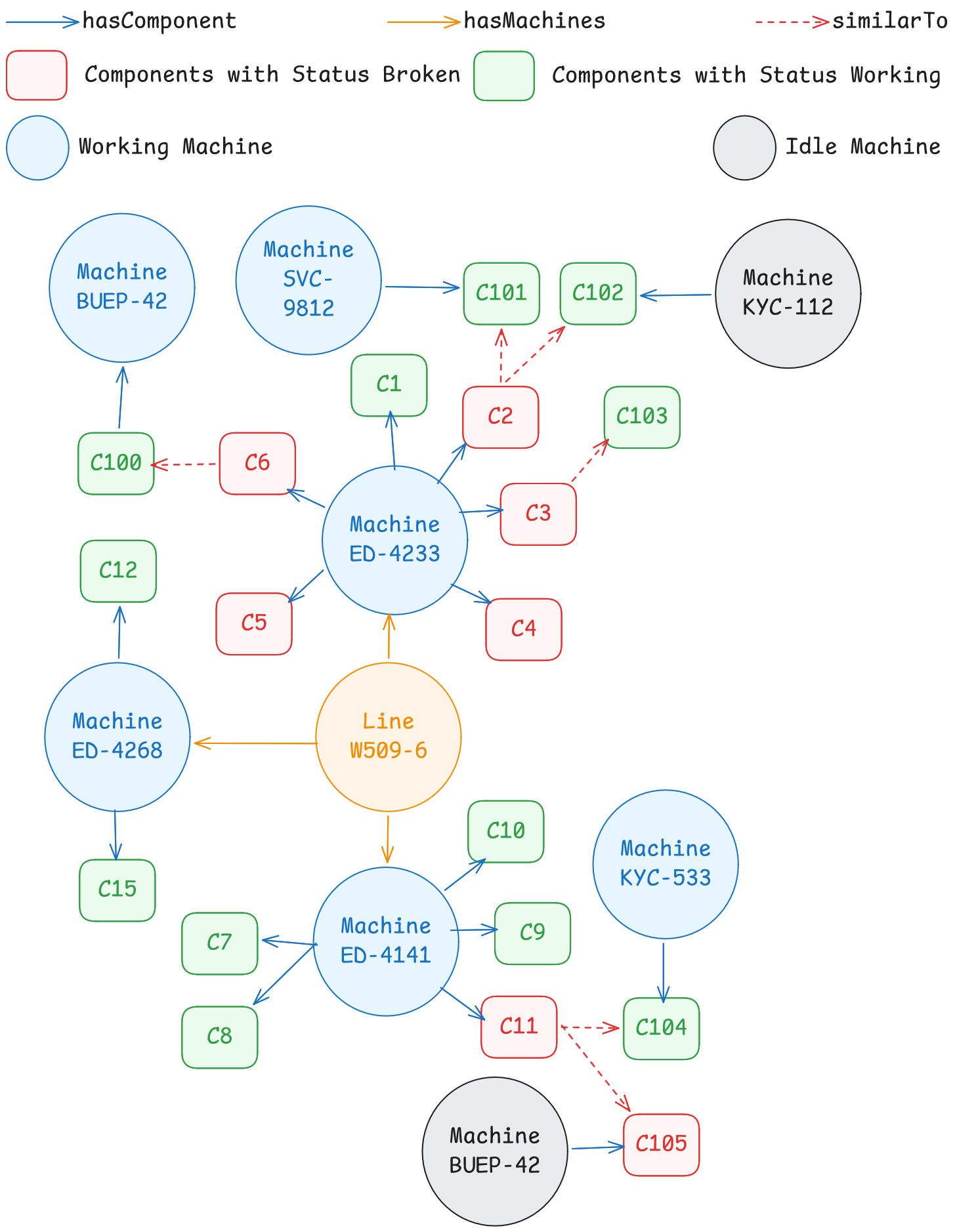}
    \caption{Representative fragment of CMDB-KG.}
    \label{fig:complex_example}
\end{figure}

\begin{example}[CMDB-KG]
Figure~\ref{fig:complex_example} shows a representative fragment of CMDB-KG. A production line (e.g., \texttt{Line W509-6}) is linked to multiple machines via \texttt{hasMachine}, and each machine connects to its installed components through \texttt{hasComponent}. Machines and components are annotated with operational status (e.g., \texttt{working}, \texttt{idle}, \texttt{broken}), and components may additionally be connected by \texttt{similarTo} relations to indicate functional interchangeability.
\end{example}

\paragraph{Question-Answer Pair Generation.}
We begin by collecting representative business-oriented information needs from the enterprise setting and abstracting them into a set of structured query templates aligned with the CMDB schema. These templates are then instantiated automatically over the CMDB-KG in a scalable manner, generating executable question–answer pairs grounded in real enterprise data. The resulting benchmark comprises 9 representative query types and a total of 19,080 questions. Detailed specifications are provided in the Appendix~\ref{app:question_construciton}.

\begin{example}[Business-Oriented Question]
    Consider the question: \emph{“Which working components can replace broken components installed on machines in production line \texttt{W509-6}?”}
    Unlike public benchmark questions, this query cannot be expressed as a single well-defined compositional logical form. It requires identifying broken components in \texttt{W509-6}, retrieving functionally equivalent components via \texttt{similarTo}, and enforcing operational constraints. In particular, valid replacements must be both working and installed on idle machines; otherwise, removing them would disrupt other production lines. These implicit business constraints must be respected during reasoning.
\end{example}

\paragraph{Evaluation Protocol.}
We follow the evaluation protocol of \citet{zhou2025evaluating, DBLP:journals/corr/abs-2508-08344}. Let $\mathcal{P}_q$ and $\mathcal{A}_q$ denote the predicted and ground-truth answer sets for question $q$, respectively. 
\emph{Accuracy} measures exact set match ($\mathcal{P}_q = \mathcal{A}_q$). 
\emph{Hits@Any} measures whether at least one correct answer is retrieved ($\mathcal{P}_q \cap \mathcal{A}_q \neq \emptyset$). 
\emph{Precision} and \emph{Recall} quantify answer correctness and completeness based on set overlap, and \emph{F1} is their harmonic mean.

\subsection{Limitations of Existing Agentic KG-RAG Paradigms}
\label{sec:failure_analysis}

Through qualitative analysis of model outputs and reasoning traces on the enterprise benchmark, we observe systematic limitations in both dominant agentic KG-RAG paradigms: plan-and-execute and ReAct-style iterative exploration.

\paragraph{Limitations of plan-and-execute approaches.}
Plan-and-execute methods mitigate uncontrolled exploration by learning explicit reasoning plans or subgraph selection strategies. However, they rely on substantial training and hyperparameter tuning to perform reliably, which is particularly challenging when adapting large models in enterprise environments. In practice, effective training requires sufficient coverage of the underlying query and relation distributions; otherwise, learned planning patterns tend to overfit to training-specific structures and fail to generalize to unseen schemas or reasoning compositions. This dependence on distribution-specific training limits robustness under evolving enterprise KGs.

\paragraph{Limitations of ReAct-style exploration.}
ReAct-based methods rely on iterative, relevance-driven traversal. In dense enterprise graphs, this often leads to \emph{search explosion}, as models expand through high-degree attribute nodes (e.g., identifiers or status values), rapidly increasing the search space. Moreover, traversal guided primarily by semantic similarity frequently results in \emph{schema-agnostic reasoning}, producing paths that are semantically plausible but structurally invalid under enterprise schema constraints. As reasoning depth increases, these issues compound into \emph{semantic drift}, where the agent deviates from the original query intent due to locally relevant yet globally irrelevant paths.

\section{SCAIR: Schema-Conditioned Agentic Iterative Reasoning}
\begin{figure}[t]
    \centering
    \includegraphics[width=\columnwidth]{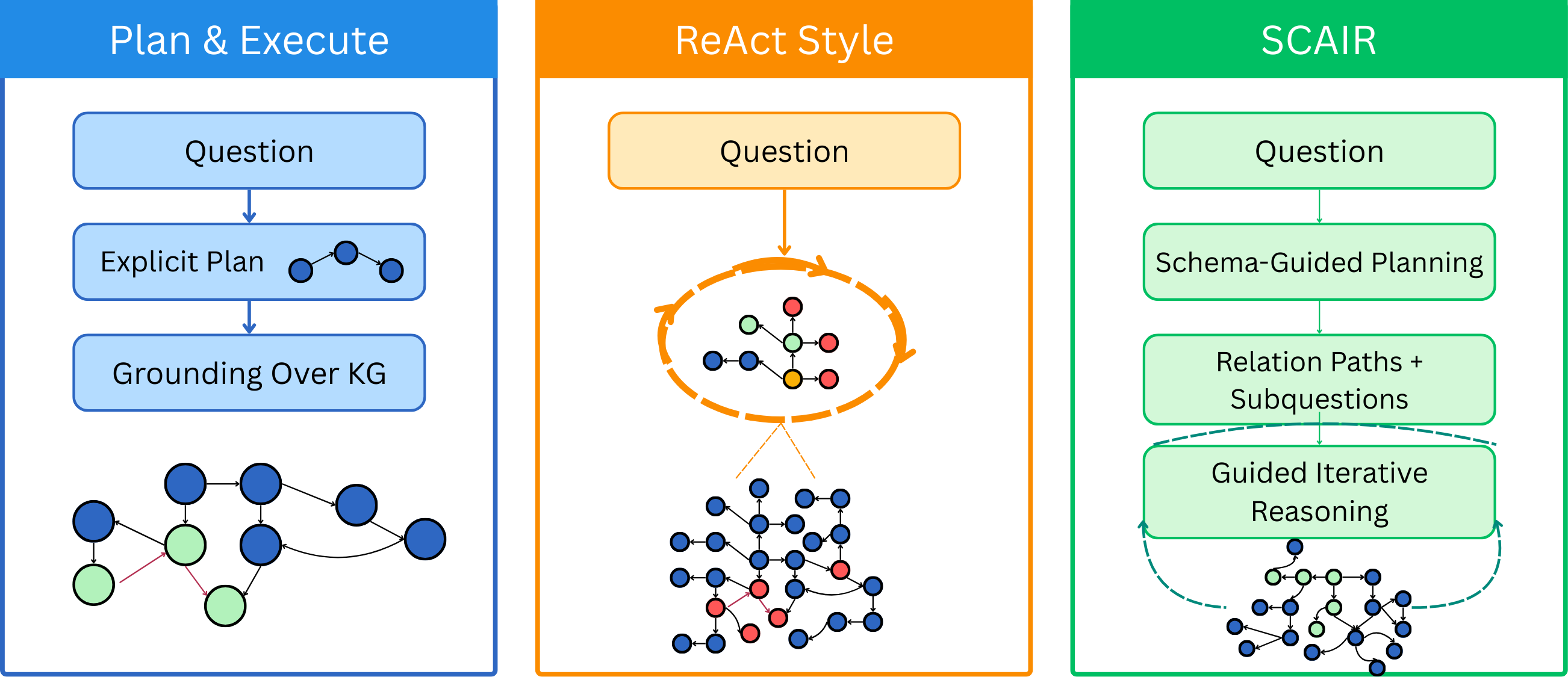}
    \caption{Overview of Plan-and-Execute, ReAct and SCAIR.}
    \label{fig:kgrag_overview}
\end{figure}

To address the enterprise-specific limitations identified in Section~\ref{sec:failure_analysis}, we propose \textbf{SCAIR}, a novel KG-RAG method built on a \textbf{training-free hybrid agentic paradigm} (Figure~\ref{fig:kgrag_overview}). Unlike existing approaches that adopt either a plan-and-execute strategy or a ReAct-style iterative exploration (Section~\ref{sec:existing_methods}), SCAIR integrates structured planning with controlled iterative reasoning within a unified framework.

SCAIR is guided by three design principles: (i) schema-conditioned planning to provide structural priors, (ii) schema-aware iterative reasoning to ensure valid and focused multi-hop traversal, and (iii) controlled topic entity propagation to balance exploration and exploitation during search.

\paragraph{Schema-Conditioned Planning.}
Before iterative reasoning (i.e. the ReAct loop), SCAIR performs a lightweight planning stage to introduce structural priors. Specifically, we generate schema-consistent relation paths based on entity types and relation definitions to restrict traversal to valid relation compositions and avoid expansion through irrelevant high-degree nodes. In parallel, the input question is decomposed into depth-aligned subquestions that specify the information required at each reasoning hop. Unlike strict plan-and-execute systems, this stage does not commit to a single fixed plan, but instead provides structural guidance that constrains subsequent exploration.

\paragraph{Schema-Aware Iterative Reasoning.}
The core of SCAIR follows an iterative agentic loop. At each depth, candidate relations connected to the current topic entities are retrieved, filtered using schema constraints, and scored conditioned on the corresponding subquestion. The selected relations are then expanded to candidate entities, which are further scored and pruned based on the accumulated reasoning context. 
Traversal is thus guided jointly by semantic relevance and schema validity, preventing structurally invalid paths and reducing search explosion in dense enterprise graphs. The process terminates once sufficient evidence is collected or the maximum reasoning depth is reached.

\paragraph{Exploitation-Exploration Tradeoff.}
SCAIR balances exploration and exploitation through controlled topic entity propagation. Rather than replacing the topic entity set at each depth, the algorithm maintains a union of previously discovered entities and newly expanded ones. This preserves earlier reasoning anchors (exploitation) while enabling deeper traversal into new regions of the graph (exploration). In contrast, ReAct-style methods typically update the working state using only newly expanded entities, which can prematurely discard earlier topic entities and limit the opportunity to explore alternative relations connected to them.

\section{Experiments and Results}

\subsection{Experimental Settings}
All methods are evaluated on the enterprise CMDB benchmark. We compare SCAIR with agentic KG-RAG paradigms: plan-and-execute methods (RoG, G-Retriever) and ReAct-style iterative methods (ToG, PoG), covering training-based and training-free strategies. Baseline model configurations and hyperparameters follow the settings reported in the original papers to ensure fair comparison. Detailed configurations are provided in Table~\ref{tab:baseline_settings} in Appendix \ref{app:baseline_details}. Training-based models are fine-tuned on two NVIDIA A100 GPUs, while training-free methods are evaluated via OpenAI API.

\subsection{Results}
\label{sec:quantitative_results}
\paragraph{Overall Performance.}
Table~\ref{tab:overall_comparison} reports the performance of baseline KG-RAG methods and SCAIR on the CMDB benchmark. Overall, baseline performance remains limited across all metrics. Among the baselines, training-based methods perform better than training-free approaches. In particular, G-Retriever achieves the strongest aggregate results among baselines (25.27 Accuracy, 24.28 F1), while RoG attains the highest Hits@Any (38.49) but with lower precision and recall. Training-free methods (ToG and PoG) consistently underperform across metrics.
Importantly, SCAIR outperforms all baselines across evaluation metrics. These results indicate that while existing KG-RAG paradigms struggle under enterprise KG conditions, aligning reasoning with schema structure, as done in SCAIR, leads to consistent and significant gains.

\begin{figure}[t]
    \centering
    \includegraphics[width=\columnwidth]{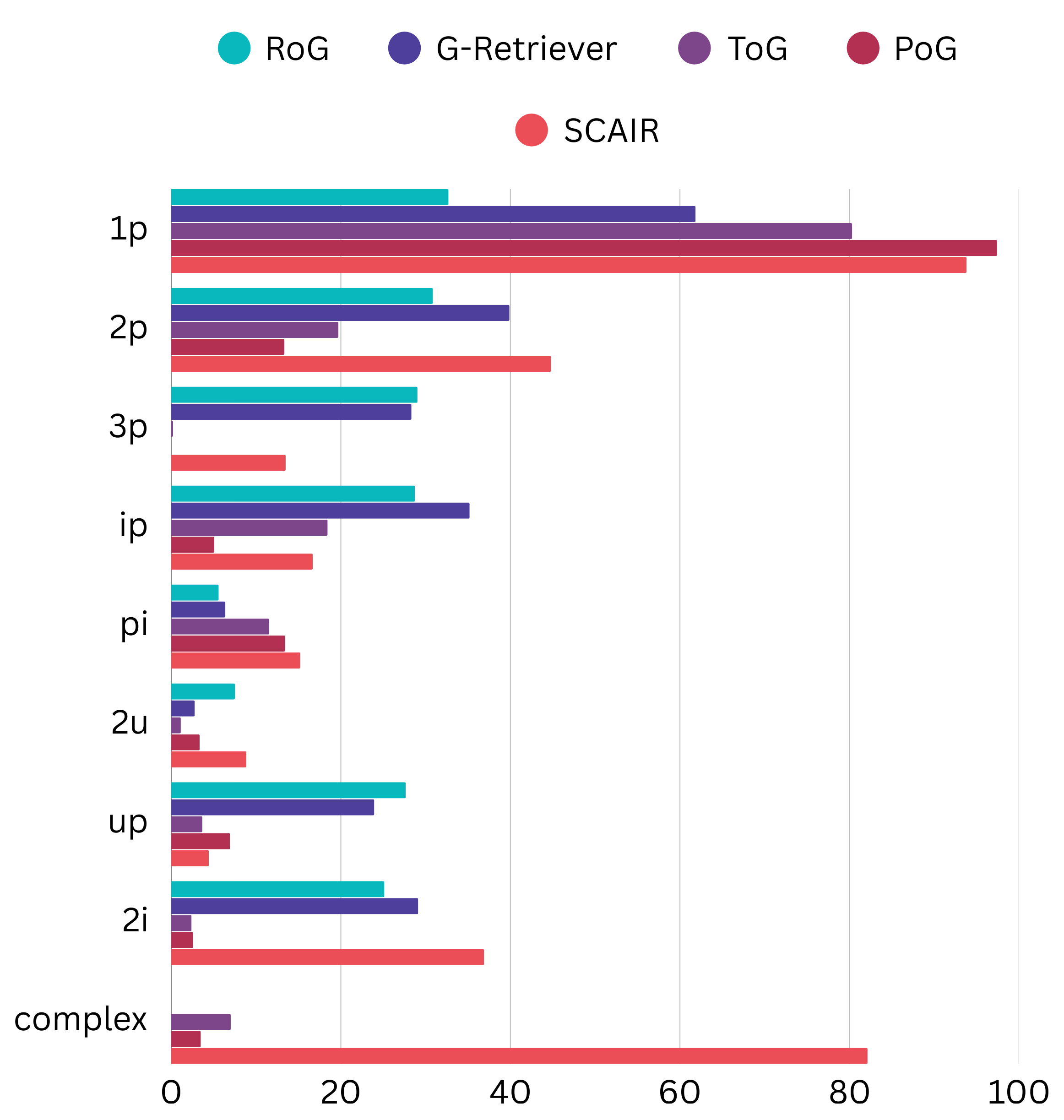}
    \caption{Accuracy comparison across query types. Results for other evaluation metrics are reported in the Appendix \ref{app:per_type_metrics}.} 
    \label{fig:improved_accuracy_per_type}
\end{figure}

\paragraph{Performance Across Query Types.}
To better understand the source of the performance gains, we analyze accuracy across query types (Figure~\ref{fig:improved_accuracy_per_type}). 
SCAIR consistently outperforms training-free ReAct-style methods (ToG and PoG) across nearly all categories, with especially large improvements on multi-hop (2p, 3p), intersection (2i), and complex queries. These categories are particularly sensitive to uncontrolled traversal and semantic drift, where ReAct-based exploration often fails to maintain structural validity.

Compared to training-based approaches (RoG and G-Retriever), SCAIR achieves competitive or superior performance across most query types. While G-Retriever performs strongly on simpler path queries (e.g., 1p), its accuracy drops substantially on complex and constraint-heavy queries. This pattern suggests that training-based models tend to internalize reasoning templates prevalent in the training distribution, but struggle to generalize. In contrast, SCAIR maintains high accuracy even in the most challenging \emph{complex} category, indicating stronger robustness to distribution shifts and structurally diverse reasoning patterns.
The same trends are reflected in Hits@Any and F1 scores (see Appendix~\ref{app:per_type_metrics}).

Overall, the improvements are distributed across query types rather than concentrated on isolated patterns, suggesting that the gains arise from more reliable control over traversal and reasoning structure rather than from stronger language modeling or memorization.

\begin{figure}[t]
    \centering
    \includegraphics[width=0.9\columnwidth]{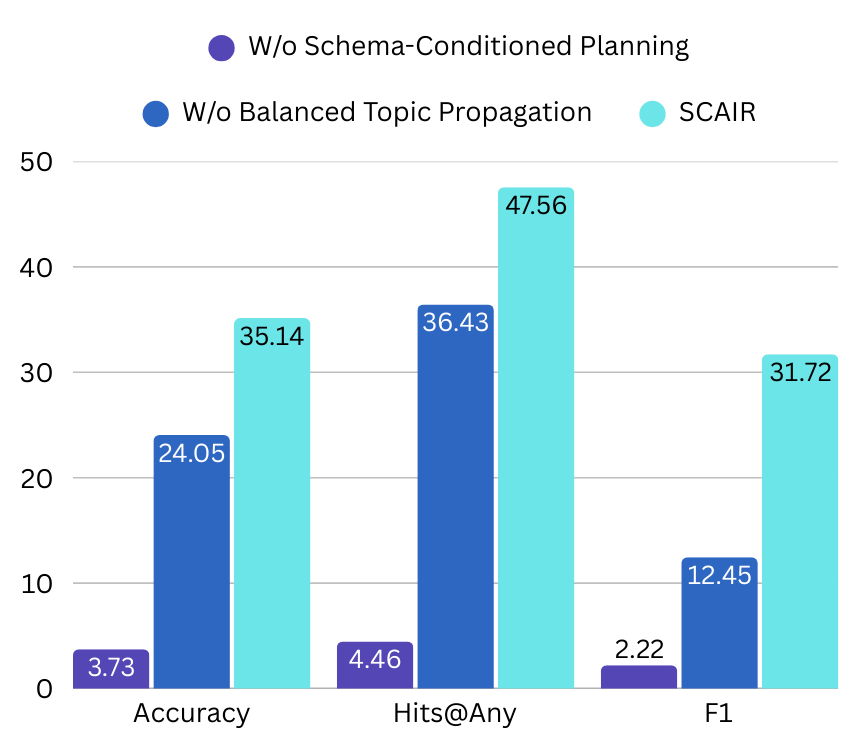}
    \caption{Component Impact on Overall Performance.}
    \label{fig:ablation_components}
\end{figure}

\paragraph{Ablation Study.}
We analyze the impact of structural control by selectively disabling schema-aware relation filtering and entity-centric constraints in the proposed
agentic KG-RAG framework. As shown in Figure~\ref{fig:ablation_components}, removing these controls leads to a consistent degradation in performance across query categories. Overall, the results indicate that explicit structural constraints play a central role in the robustness of the proposed approach.

\begin{table}[t]
\centering
\resizebox{0.49\textwidth}{!}{%
\begin{tabular}{lccccc}
\hline
\textbf{Method} & \textbf{Accuracy} & \textbf{Hits@Any} & \textbf{F1} & \textbf{Precision} & \textbf{Recall} \\
\hline
RoG          & 20.80 & 38.49 & 19.09 & 23.32 & 20.37 \\
G-Retriever & 25.27 & 36.60 & 24.28 & 28.25 & 24.95 \\
ToG & 16.04 & 20.26 & 14.65 & 15.62 & 15.07 \\
PoG         & 16.17 & 21.41 & 14.49 & 15.50 & 14.60 \\
\hline
\textbf{SCAIR} & \textbf{35.14} & \textbf{47.56} & \textbf{31.72} & \textbf{36.78} & \textbf{32.60} \\
\hline
\end{tabular}
}
\caption{Overall performance comparison.}
\label{tab:overall_comparison}
\end{table}

\paragraph{Inference Cost Analysis.}
We report per-query inference cost for training-free baselines and SCAIR in Table \ref{tab:inference_cost}.
SCAIR requires more LLM calls (42.36) and tokens (22.5k input, 5.9k output) than ToG (13.95; 5.9k/2.1k) and PoG (5.60; 3.7k/0.4k), reflecting its deeper average traversal (2.54 vs. 2.14/1.61).
This cost is functional. Most benchmark query types (i.e. 2p, 3p, ip, pi, and complex) require 2–4 reasoning hops, and SCAIR's depth of 2.54 approximates the minimum needed for correct multi-hop resolution. ToG and PoG terminate earlier and consequently underperform on these categories. 

Token growth is input-dominated, arising from accumulating reasoning context across hops; this structure is particularly amenable to API prompt caching, where repeated prefixes receive 50–90\% cost discounts \cite{openai2025promptcachingdocs} and substantially reduce effective per-query cost in deployed settings. Direct cost comparison with training-based methods is not meaningful, as their total cost includes supervision curation and GPU training time not captured by inference metrics. SCAIR is therefore suited to enterprise workloads where answer correctness is prioritized over single-query latency.

\begin{table}[t]
\centering
\resizebox{0.49\textwidth}{!}{%
\begin{tabular}{lcccccc}
\toprule
\textbf{Method} & \textbf{Calls/Q} & \textbf{Input Tok} & \textbf{Output Tok} & \textbf{Depth} \\
\midrule
ToG   & 13.95 & 5858  & 2081 & 2.14 \\
PoG   & 5.60  & 3699  & 410  & 1.61 \\
SCAIR & 42.36 & 22461 & 5931 & 2.54 \\
\bottomrule
\end{tabular}
}
\caption{Per-query inference cost for training-free agentic KG-RAG methods. Calls/Q denotes average LLM calls per question; Depth denotes average reasoning depth reached.}
\label{tab:inference_cost}
\end{table}

\section{Discussion}

\subsection{Implications for Industrial Deployment}
Our results challenge the implicit assumption in KGQA literature that improved language modeling automatically translates to deployment readiness. In enterprise settings, the "correctness" of an answer is bound by implicit operational constraints (e.g., component availability or compatibility) rather than just semantic relevance. We identify three critical principles for deploying KG-RAG in such dense, schema-governed environments:

\paragraph{1. Structural Validity Must Gate Semantic Relevance.}
In dense CMDBs, semantic similarity is a noisy proxy for utility. High-degree attribute nodes (e.g., "Status: Broken") often act as "semantic supernodes," causing ReAct-style agents to drift into operationally irrelevant subgraphs. Effective retrieval must therefore be \textit{structure-first}: schema constraints should prune the search space \textit{before} semantic scoring occurs, preventing the "hallucinated validity" observed in standard baselines.

\paragraph{2. Traversal Control is the Primary Bottleneck.}
Comparing LLaMA-2 and Qwen-2.5 backbones (Appendix \ref{app:qwen_experiment}) reveals that stronger parametric knowledge improves answer generation but fails to prevent search explosion. The failure mode is architectural, not parametric. For practitioners, this implies that investing in lightweight, schema-aware planning yields higher reliability gains than simply scaling the inference backbone.

\paragraph{3. Inference-Time Adaptation Outperforms Retraining.}
Plan-and-execute methods often overfit to specific query templates seen during training, making them brittle to the frequent schema evolutions typical of enterprise IT. Training-free frameworks like SCAIR, which inject structural priors at inference time, offer a more maintainable deployment strategy. They allow the reasoning engine to adapt to new business rules or schema updates without the cost of continuous supervised fine-tuning.

\section{Conclusion and Future Work}
This work exposes the fragility of current agentic KG-RAG paradigms when applied to the dense, constraint-heavy reality of industrial KGs. By introducing a realistic CMDB benchmark, we demonstrate that dominant failure modes stem from a misalignment between open-domain retrieval heuristics and enterprise operational logic.

Our approach SCAIR mitigates these issues by enforcing
structural alignment within the agentic loop, but the path to
fully autonomous enterprise agents requires further evolution
along several directions. First, moving beyond prompt-based
heuristics, business rules such as disruption risk and valid
replacement conditions could be encoded as explicit schema
annotations or typed constraints, enabling more principled
reasoning over operational logic. Second, while inference-time
control is effective, designing training objectives that reward
structural validity rather than only final-answer accuracy
remains a promising direction for next-generation graph
reasoning models. Finally, extending the empirical comparison
to non-agentic schema-constrained traversal and text-to-SPARQL
baselines would further disentangle the contributions of
structural filtering from agentic iteration.

\section{Acknowledgements}
The authors thank the International Max Planck Research School for Intelligent Systems (IMPRS-IS) for supporting Yuqicheng Zhu, Hongkuan Zhou and Yunjie He. 
The work was partially supported by EU Projects SMARTY (GA 101140087).



\newpage
\bibliography{anthology}

\clearpage
\appendix

\section{Benchmark construction details}
\label{app:dataset_construction}

\subsection{CMDB-KG Construction}
\label{app:kg_construciton}

\begin{figure}[t]
    \centering
    \includegraphics[width=1\linewidth]{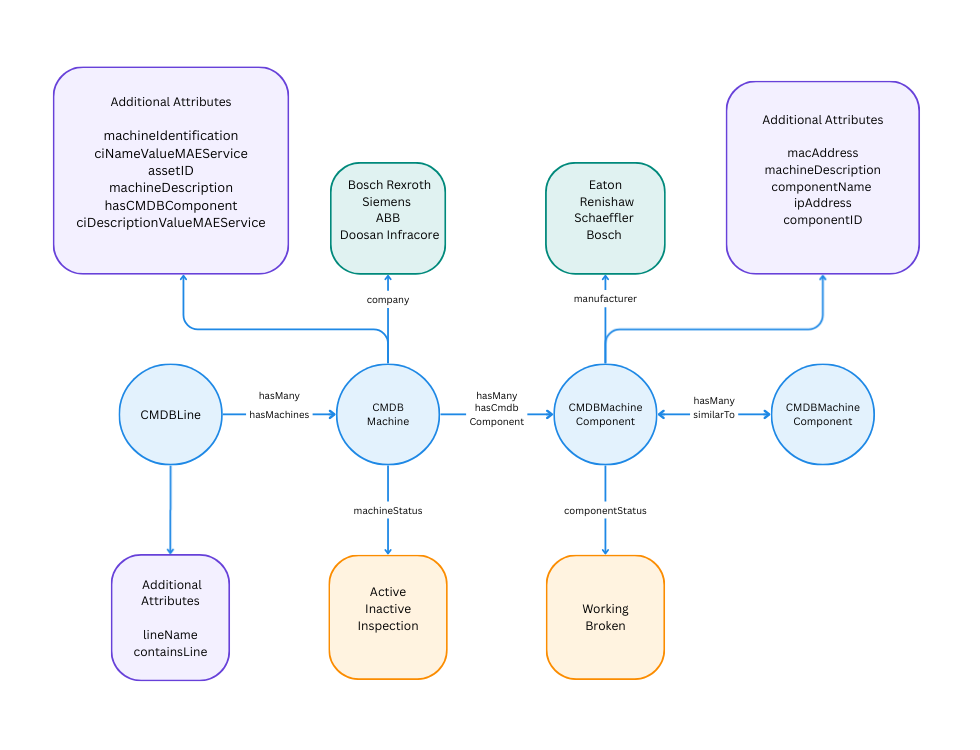}
    \caption{High-level schema overview of the enterprise CMDB KG. The figure illustrates the core entity types (lines, machines, components), their relations, and selected operational attributes introduced to support enterprise reasoning tasks.}
    \label{fig:cmdb_overview}
\end{figure}

\paragraph{KG Enhancement.} 
To support realistic reasoning patterns, we extend the original CMDB KG with a small set of auxiliary relations commonly required in operational scenarios, such as component status, manufacturer information, and functional similarity between components. These extensions do not alter the core semantics of the CMDB but enable queries related to diagnostics, filtering, and component replacement that are central to enterprise use cases. (Figure~\ref{fig:cmdb_overview}) illustrates the high-level schema structure of the CMDB KG which supports complex diagnostic and replacement queries.

\subsection{Question-Answer Pair Construction.}
\label{app:question_construciton}

\begin{figure}[t]
    \centering
    \includegraphics[width=1\linewidth]{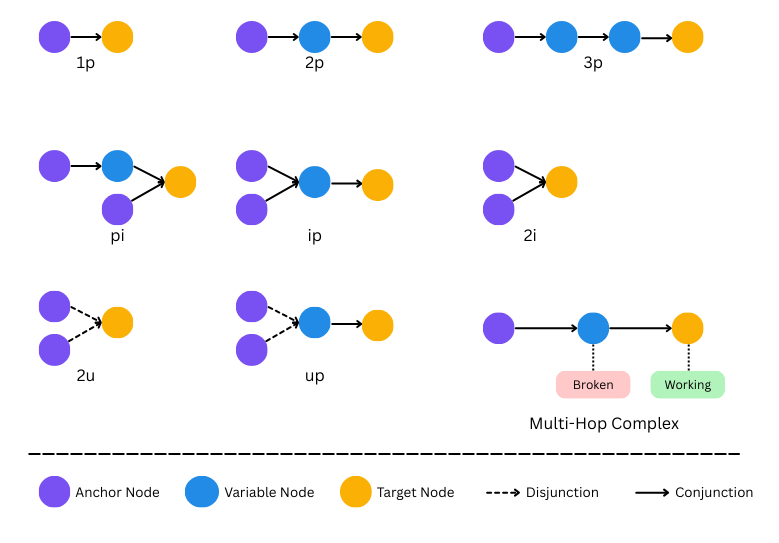}
    \caption{Illustrative examples of structured reasoning templates used in this work. The visualization
style and template taxonomy are inspired by \citep{DBLP:conf/nips/ArakelyanMDCA23}}
    \label{fig:query_types}
\end{figure}

Rather than collecting natural language questions from crowd workers or logs, we construct the benchmark using a template-based question generation framework grounded in the CMDB schema. 
Each question template corresponds to a predefined reasoning pattern (illustrated in Figure~\ref{fig:query_types}), including multi-hop projection, logical conjunction, disjunction, and domain-specific compositional reasoning. The benchmark covers nine query categories: single-hop projection (1p), multi-hop projection (2p, 3p), intersection (2i), path–intersection hybrids (ip, pi), union (2u, up), and domain-specific complex queries involving operational constraints.

\begin{figure}[t]
\centering
\includegraphics[width=0.9\linewidth]{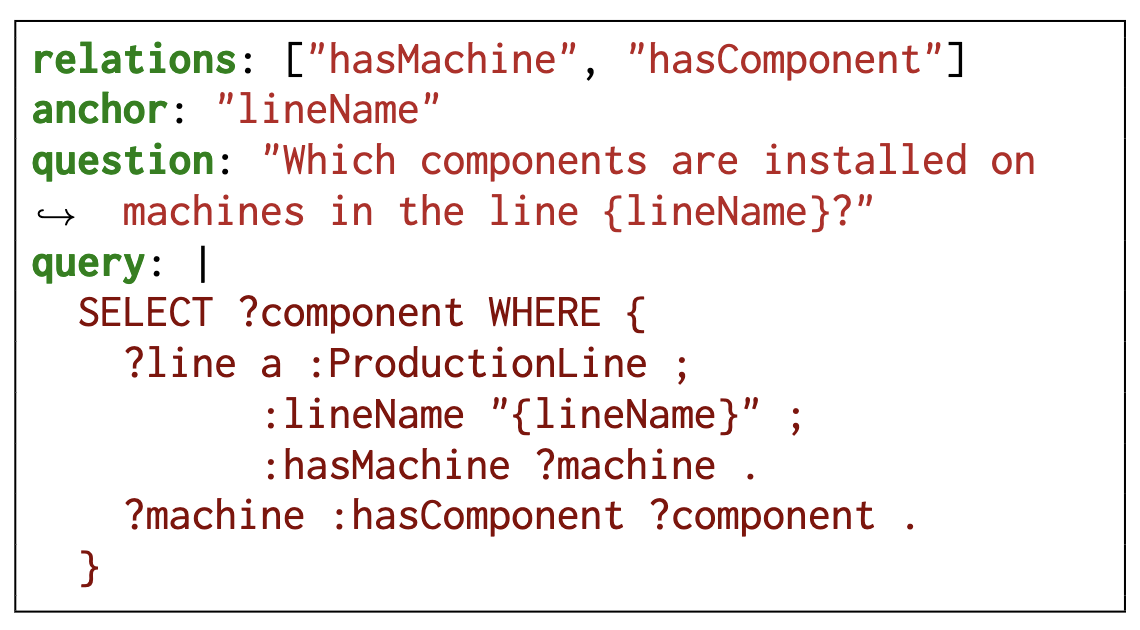}
\caption{Example of a template-based question specification and its executable
graph query used for benchmark construction.}
\label{fig:template_example}
\end{figure}

Templates are defined over the CMDB schema and specify both the underlying graph traversal pattern and a natural language realization. During instantiation, template placeholders are grounded with concrete entities from the CMDB KG, and the resulting queries are executed against the KG to obtain gold-standard answers. An example of a template specification and its corresponding executable query is shown in Figure~\ref{fig:template_example}.

In addition to standard compositional patterns, the benchmark includes domain-specific multi-hop queries that reflect real industrial scenarios. For example, replacement queries require identifying functionally similar components that satisfy operational constraints, such as being currently functional while the original component is broken. These queries combine topological traversal with attribute-based filtering and implicit business logic, making them particularly challenging for KG-RAG systems.

As shown in Table~\ref{tab:eval_dataset_query_templates}, the benchmark contains 19,080 questions in total. While single-hop (1p) queries dominate in absolute number due to large entity coverage, the benchmark includes 7,080 compositional queries (excluding 1p), spanning multi-hop, intersection, union, and domain-specific complex patterns. This distribution reflects realistic enterprise workloads, where simple lookups are frequent but structurally complex queries are critical for diagnostic and operational tasks.

\begin{table}[t]
\centering
\small
\setlength{\tabcolsep}{4pt}
\caption{Distribution of query types and instantiated question instances.}
\label{tab:eval_dataset_query_templates}
\begin{tabular}{lrr}
\toprule
\textbf{Type} & \textbf{\#Temp.} & \textbf{\#Inst.} \\
\midrule
1p & 1  & 12,000 \\
2p & 22 & 1,690  \\
3p & 14 & 930    \\
2i & 27 & 1,497  \\
ip & 14 & 590    \\
pi & 18 & 1,202  \\
2u & 12 & 516    \\
up & 12 & 577    \\
complex & 1 & 78 \\
\midrule
\textbf{Complex (excl. 1p)} & -- & 7,080 \\
\textbf{Total} & -- & 19,080 \\
\bottomrule
\end{tabular}
\end{table}

\section{Case Study of Existing Methods}
\label{app:case_study}
We report representative failure cases observed for ReAct style ToG on the enterprise benchmark. Each case includes (i) a real benchmark question, (ii) the intended KG reasoning pattern, and (iii) a concrete failure symptom consistent with the failure modes.

\subsection{Search Explosion}
\paragraph{Question (complex).}
\emph{Which working components can replace broken components installed on machines in production line \texttt{W509-6}?}

\paragraph{Intended KG reasoning.}
\texttt{Line W509-6} $\xrightarrow{\texttt{hasMachine}}$ machines
$\xrightarrow{\texttt{hasComponent}}$ components
$\xrightarrow{\texttt{componentStatus}}$ filter \texttt{broken}
$\xrightarrow{\texttt{similarTo}}$ candidates
$\xrightarrow{\texttt{componentStatus}}$ filter \texttt{working}.
Additionally, candidates installed on \texttt{idle} machines are preferred to avoid disrupting other lines.

\paragraph{Observed failure symptom.}
ReAct-style exploration frequently expands via high-degree attribute relations before committing to \texttt{similarTo}. A typical trajectory is:
\begin{quote}\small
\textbf{Step 1:} retrieve machines for \texttt{W509-6} (\texttt{hasMachine}) \\
\textbf{Step 2:} retrieve components (\texttt{hasComponent}) \\
\textbf{Step 3:} expand via \texttt{manufacturer} or \texttt{componentStatus} (hub attribute) \\
\textbf{Step 4:} retrieve many unrelated ``working'' components sharing the same attribute value
\end{quote}
This produces large candidate sets dominated by generic attribute matches (e.g., many components linked to \texttt{working}), exhausting the beam budget and preventing exploration along the intended \texttt{similarTo} relation.

\paragraph{Example incorrect output.}
Predicted answer sets often contain many unrelated working components (high Hits@Any but low precision), e.g.,
\{\texttt{P-E11-26838}, \texttt{P-E11-27317}, \texttt{P-E11-31641}, \dots\},
where the returned components share an attribute hub (e.g., \texttt{working}) but are not valid replacements for the broken components in \texttt{W509-6}.

\subsection{Schema-Agnostic Reasoning}
\paragraph{Question (3p-style).}
\emph{Which IP addresses are assigned to components installed on machines in production line W509-6?}

\paragraph{Intended KG reasoning.}

\texttt{ProductionLine}
$\xrightarrow{\texttt{hasMachine}}$
\texttt{Machine}
$\xrightarrow{\texttt{hasComponent}}$
\texttt{Component}
$\xrightarrow{\texttt{ipAddress}}$
\texttt{IP}.

\paragraph{Observed incorrect traversal.}
Instead of expanding toward \texttt{ipAddress}, the agent may select a semantically plausible relation such as \texttt{manufacturer}:

\texttt{Machine}
$\xrightarrow{\texttt{manufacturer}}$
\texttt{BoschX}
$\xrightarrow{\texttt{manufacturer}^{-1}}$
\texttt{Machine}.

Because network configuration is often associated with vendors in real-world settings, this transition appears relevant at the language level. However, it expands through a high-degree manufacturer node and retrieves machines unrelated to the queried production line.

\subsection{Semantic Drift Under Increasing Reasoning Depth}

\paragraph{Question (3p-style).}
\emph{Which production lines contain machines that use components manufactured by Siemens?}

\paragraph{Intended KG reasoning.}

\texttt{Manufacturer(Siemens)}
$\xrightarrow{\texttt{manufacturer}^{-1}}$
\texttt{Component}
$\xrightarrow{\texttt{hasComponent}^{-1}}$
\texttt{Machine}
$\xrightarrow{\texttt{hasMachine}^{-1}}$
\texttt{ProductionLine}.

\paragraph{Observed drift pattern.}
After reaching \texttt{Component} nodes, the agent may select a semantically plausible but task-irrelevant relation such as \texttt{similarTo}:

\texttt{Component}
$\xrightarrow{\texttt{similarTo}}$
\texttt{Component}
$\xrightarrow{\texttt{hasComponent}^{-1}}$
\texttt{Machine}.

Because similarity is often associated with compatibility, this transition appears reasonable at the language level. However, it expands the search to components not manufactured by Siemens, thereby violating the original constraint.

\paragraph{Effect.}
Although each hop is schema-valid, the reasoning progressively drifts from the manufacturer constraint. As depth increases, locally plausible transitions accumulate, leading to production lines unrelated to Siemens-manufactured components. This illustrates semantic drift in multi-hop enterprise reasoning.

\paragraph{Summary.}
These cases instantiate the failure modes discussed in Section~\ref{sec:failure_analysis}: uncontrolled expansion via high-degree hubs, schema-agnostic relation selection, and progressive semantic drift under increasing reasoning depth.

\section{More Details of SCAIR}
\label{app:scair_details}

Algorithm~\ref{alg:improved_tog} provides the full pseudocode of SCAIR. In the following sections, we also detail implementation parameters, prompting strategy, 
and search control mechanisms used in our experiments.

\label{app:algorithm}
\begin{algorithm}[t]
\caption{Schema-Aware and Question-Guided Method}
\label{alg:improved_tog}
\begin{algorithmic}[1]
\Require Question $Q$, initial topic entities $E_0$, beam width $w$, maximum depth $d$, KG $\mathcal{G}$
\Ensure Answer set $\mathcal{P}_Q$

\State $\mathcal{P} \leftarrow \Call{GenerateRelationPaths}{Q, \mathcal{G}}$
\Comment{schema-guided relation paths}

\State $\{q_1, q_2, \dots, q_d\} \leftarrow \Call{DecomposeQuestion}{Q, \mathcal{P}}$
\Comment{depth-aligned subquestions}

\State $E_0 \leftarrow$ given topic entities

\For{$t = 0$ \textbf{to} $d-1$}

    \State \textbf{Relation Search:}
    \State $\mathcal{R}_t \leftarrow \Call{RetrieveRelations}{E_t, \mathcal{G}}$
    \State $\mathcal{R}_t \leftarrow \Call{FilterRelations}{\mathcal{R}_t, q_{t+1}, \mathcal{P}}$
    \State Score $\mathcal{R}_t$ using LLM$(Q, q_{t+1}, E_t)$
    \State Refine scores using relation paths $\mathcal{P}$
    \State $\mathcal{R}_t \leftarrow \Call{TopK}{\mathcal{R}_t, w}$

    \State \textbf{Entity Search:}
    \State $E_t' \leftarrow \emptyset$
    \ForAll{$r \in \mathcal{R}_t$}
        \State $C \leftarrow \Call{RetrieveEntities}{E_t, r, \mathcal{G}}$
        \State Score $C$ using LLM$(Q, q_{t+1}, r, C, \mathcal{P})$
        \State Update search history with $(E_t, r, C)$
        \State $E_t' \leftarrow E_t' \cup C$
    \EndFor

    \If{$E_t' = \emptyset$}
        \State \textbf{Half-stop} and skip to next depth
    \EndIf

    \State $E_t' \leftarrow \Call{TopK}{E_t', w}$

    \State \textbf{Reasoning Check:}
    \If{\Call{SufficientEvidence}{$Q, E_t'$}}
        \State \Return $\Call{GenerateAnswer}{Q, E_t'}$
    \EndIf

    \State $E_{t+1} \leftarrow E_t \cup E_t'$
    \Comment{balanced topic entity propagation}

\EndFor

\State \Return $\Call{BestCandidate}{Q, E_d}$
\Comment{final half-stop}
\end{algorithmic}
\end{algorithm}

\paragraph{Search Control Mechanisms.}
To stabilize traversal in dense graphs, we apply three controls:
(i) \emph{subsampling} when relation expansion yields more than 20 entity candidates,
(ii) \emph{cycle prevention} by tracking visited (entity, relation) expansions and discarding repeats,
and (iii) a \emph{half-stop} policy: if no valid entity remains at depth $t$, we skip expansion and continue with the previous topic state at depth $t{+}1$.

\subsection{Reasoning Parameters and Execution Setup}

SCAIR is implemented using GPT-4.1-mini via API-based inference. We adopt a multi-turn chat format with explicit reasoning instructions.

Decoding parameters are set to temperature = 0.4 during exploration steps (relation and entity scoring) to allow controlled diversity, and temperature = 0.0 during reasoning and final answer generation to ensure deterministic outputs.

The maximum reasoning depth is set to $d = 4$, and beam width to $w = 6$. When entity expansion yields more than 20 candidates, we apply subsampling 
and retain the top 8 candidates per relation (pre-pruning), before beam pruning.

All questions are processed sequentially due to the multi-step API interaction. API retries are enabled to mitigate transient failures.

The choice of beam width $w = 6$ reflects a trade-off between search coverage and stability in dense enterprise graphs. Smaller beam widths restrict 
exploration and reduce recall, while larger values increase noise and semantic drift. Empirical sensitivity analysis over beam width is provided in Appendix~\ref{app:beam_sweep}, where we show that $w = 6$ yields the best balance between F1 and computational efficiency.

\subsection{Prompting Design}
\label{app:scair_prompts}

SCAIR operates entirely at inference time using structured prompts
that guide schema-conditioned planning, iterative traversal,
and answer verification.
To ensure reproducibility, we provide the core prompt templates
used in each stage below.

\paragraph{Relation Path Generation.}
Given the enterprise KG schema and the input question, the model first generates valid relation paths that are consistent with the ontology.
These paths restrict traversal to schema-valid compositions and prevent structurally invalid expansions.

\begin{tcolorbox}[title=Relation Path Generation Prompt]
You are a reasoning assistant over an industrial Knowledge Graph.

Given a question and the KG schema below, generate a valid relation path required to answer the question. Use only relations present in the schema. Assume the topic entity is already known.

KG Structure:
CMDBLine -hasMachines-> CMDBMachine -hasCmdbComponent-> CMDBMachineComponent -similarTo-> CMDBMachineComponent

Relations:
- CMDBLine: lineName, hasMachines
- CMDBMachine: machineIdentification, machineDescription, hasCmdbComponent, company, machineStatus
- CMDBMachineComponent: componentId, componentName, macAddress, ipAddress, manufacturer, componentStatus, similarTo

Output a numbered list of relation paths.

Q: <Question>
\end{tcolorbox}

\paragraph{Subquestion Decomposition.}
Conditioned on the generated relation path, the model decomposes the original question into depth-aligned subquestions. Each subquestion corresponds to a single hop in the relation path.

\begin{tcolorbox}[title=Subquestion Decomposition Prompt]
You are given:
- A question
- A relation path
- The topic entity

Decompose the question into small subquestions,
each corresponding to one relation in the path.

Rules:
- Each subquestion must be self-contained.
- Highlight the required relation using {curly braces}.
- Follow the order of the relation path exactly.
- Output only a numbered list.

Q: <Question>
Relation Path: <Path>
Topic Entity: <Entity>
\end{tcolorbox}

\paragraph{Relation Scoring.}
At each depth, candidate relations connected to the current topic entities are scored conditioned on the current subquestion. Schema-consistent relations are prioritized.

\begin{tcolorbox}[title=Relation Scoring Prompt]
Given:
- The question
- The current subquestion
- Candidate relations
- The relation path

Assign scores to relations such that:
- Relations highlighted in {curly braces} receive the highest score.
- Scores must sum to 1.
- Only consider the current subquestion (no future hops).

Q: <Question>
Sub Question: <Subquestion>
Topic Entity: <Entity>
Candidate Relations: <Relations>
Relation Path: <Path>
\end{tcolorbox}

\paragraph{Entity Scoring.}
For each selected relation, candidate entities are scored based on their usefulness for answering the current subquestion. Binary scoring (0 or 1) is used to reduce ambiguity.

\begin{tcolorbox}[title=Entity Scoring Prompt]
You are given:
- The question
- The current subquestion
- The current relation
- Candidate entities
- The relation path

Score each entity strictly for this hop:
- 1 if critical for the subquestion.
- 0 otherwise.
- Do not hallucinate entities.

Q: <Question>
Sub Question: <Subquestion>
Relation: <Relation>
Entities: <Entities>
Relation Path: <Path>
\end{tcolorbox}

\paragraph{Sufficiency Evaluation.}
Before answer generation, the model evaluates whether the retrieved triples are sufficient to answer the question.

\begin{tcolorbox}[title=Sufficiency Check Prompt]
Given a question and retrieved knowledge triples,
determine whether the information is sufficient to answer.

Output:
- {Yes} followed by the final answer in {curly braces}, or
- {No} with a brief explanation.

Q: <Question>
Knowledge Triples: <Triples>
\end{tcolorbox}

\paragraph{Answer Generation.}
If sufficient evidence is available, the model generates the final answer grounded strictly in the retrieved triples.

\begin{tcolorbox}[title=Answer Generation Prompt]
Given the question and retrieved knowledge triples,
generate the final answer.
The answer entities must be wrapped in {curly braces}.
Do not hallucinate additional entities.

Q: <Question>
Knowledge Triples: <Triples>
\end{tcolorbox}

\section{Baseline Implementation Details}
\label{app:baseline_details}

All baseline implementations follow the original architectural designs
and training procedures described in their respective publications.
We adopt the reported hyperparameters directly (see Table~\ref{tab:baseline_settings}). If a parameter is unspecified, we use default values from the official codebases.

\begin{table}[t]
\centering
\small
\setlength{\tabcolsep}{4pt}
\caption{Key configuration settings for baseline models.}
\label{tab:baseline_settings}
\begin{tabular}{l p{6.8cm}}
\toprule
\textbf{Model} & \textbf{Key Settings} \\
\midrule
RoG &
Backbone: LLaMA-2-7B-Chat (LoRA $r=8$); 
Epochs: 3; Batch: 2; LR: $2e^{-5}$; 
Optimizer: AdamW; Cosine schedule; bf16 \\

G-Retriever &
Backbone: LLaMA-2-7B; 
GNN layers: 4; Hidden: 1024; 
Epochs: 10 (early stop=2); LR: $1e^{-5}$; fp16 \\

ToG &
Model: GPT-4.1-mini; 
Max depth: 3; Beam: 3; 
Entity subsampling: 20$\rightarrow$5; 
Temp: 0.4 (explore), 0.0 (reason) \\

PoG &
Model: GPT-4.1-mini; 
Planning depth: 4; 
Relation pruning enabled; 
Temp: 0.3 (planning/reasoning) \\

\bottomrule
\end{tabular}
\end{table}

\paragraph{Hardware.}
Training-based baselines were fine-tuned on two NVIDIA A100 GPUs (80GB).
Inference-only methods were executed sequentially via API
due to their multi-step reasoning structure.
All implementations use PyTorch with mixed precision where applicable.

\section{More Experimental Results}
\label{app:per_type_metrics}

\subsection{Hits@Any and F1 Across Query Types}
This appendix reports Hits@Any and F1-score across query types for all evaluated methods. These results complement the accuracy-based analysis presented in the main text (Section~\ref{sec:quantitative_results}) and provide additional insight into partial answer overlap and retrieval robustness.

Consistent with the accuracy trends discussed in the main body, SCAIR maintains strong performance across multi-hop, intersection, and complex query categories. The improvements are reflected not only in exact-match accuracy but also in overlap-based metrics, indicating more reliable retrieval and reasoning behavior across structurally diverse query types.

\begin{figure}[t]
    \centering
    \includegraphics[width=\columnwidth]{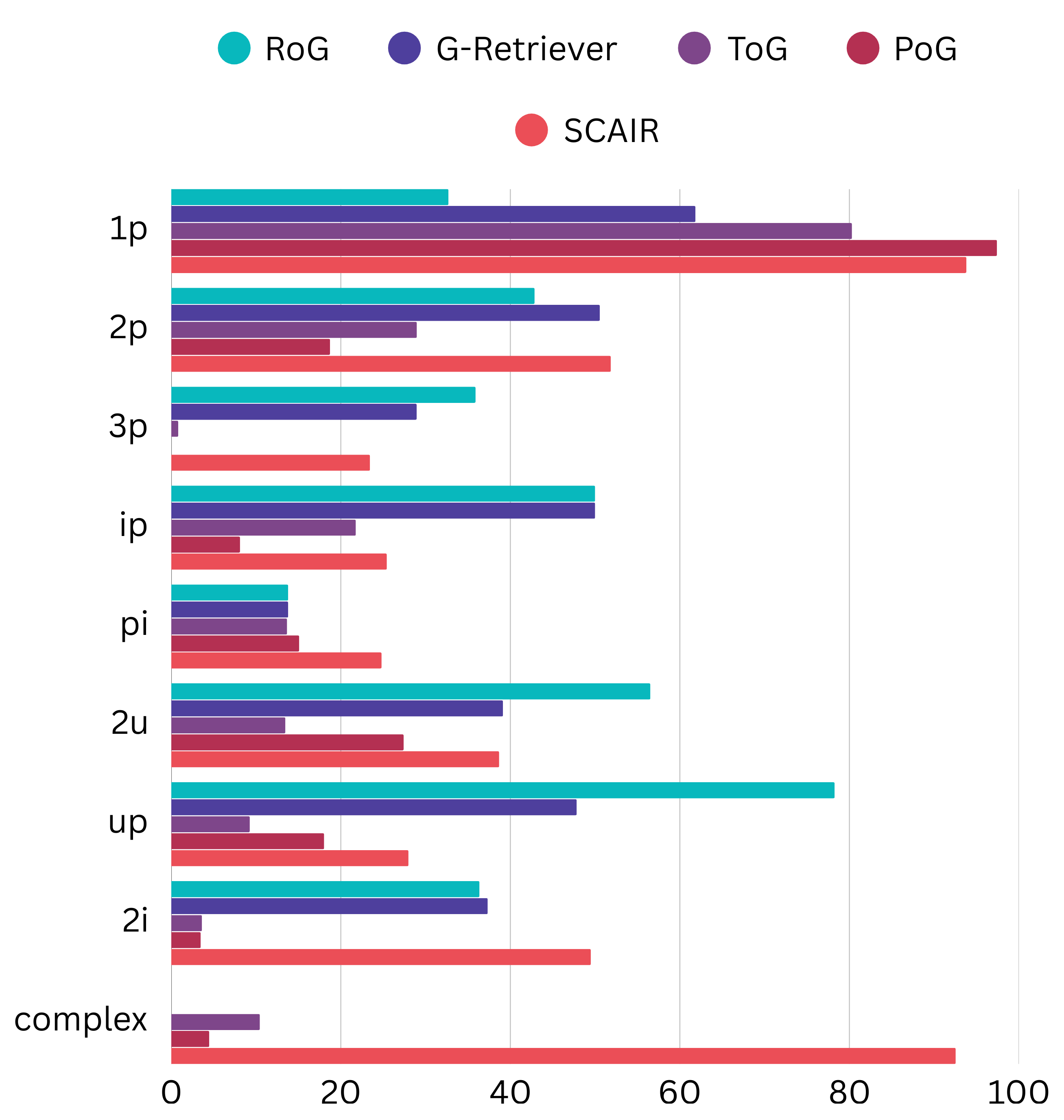}
    \caption{Hits@Any per query type for all evaluated methods.}
    \label{fig:hits_per_type}
\end{figure}

\begin{figure}[t]
    \centering
    \includegraphics[width=\columnwidth]{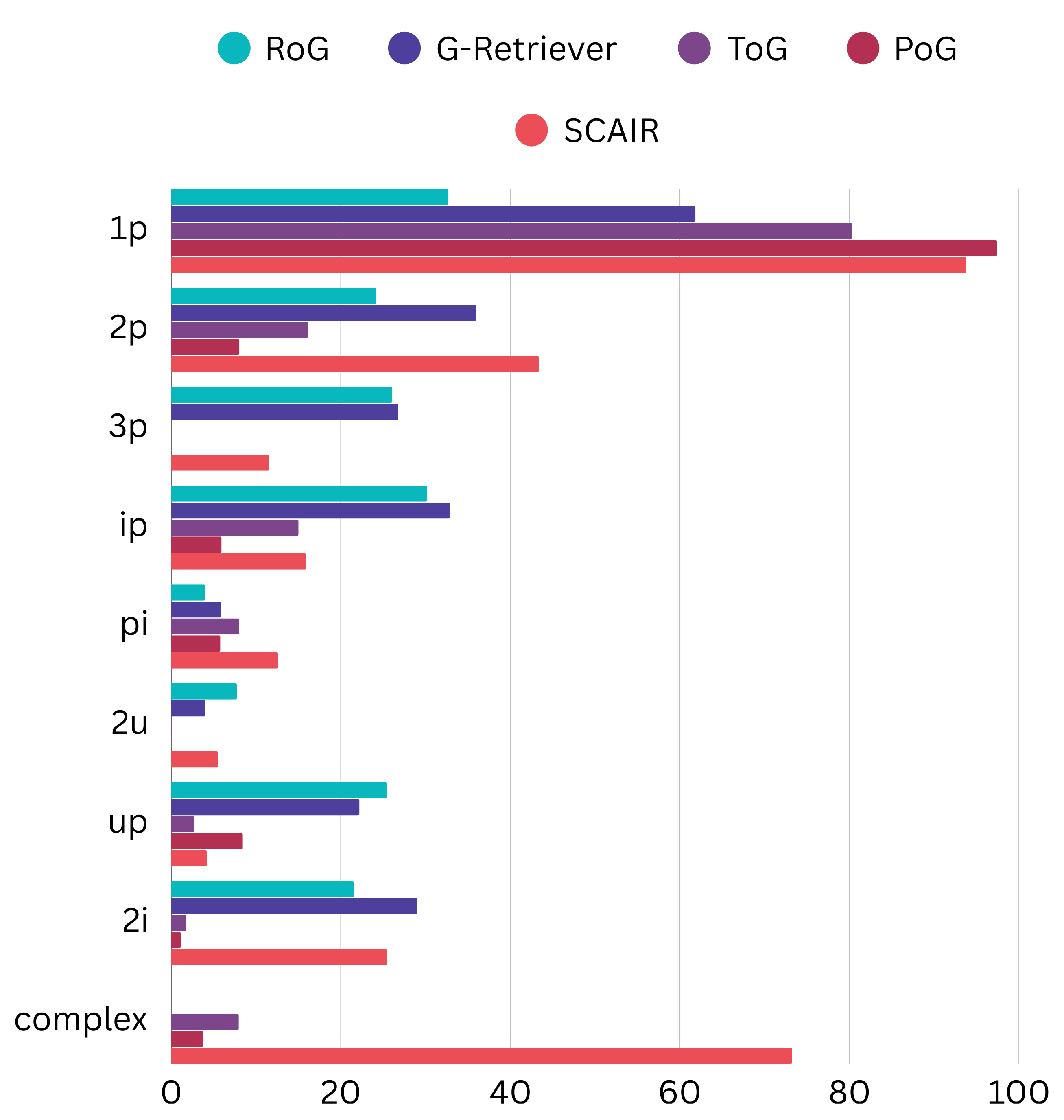}
    \caption{F1-score per query type for all evaluated methods.}
    \label{fig:f1_per_type}
\end{figure}

\subsection{Sensitivity to Beam Width}
\label{app:beam_sweep}

To analyze the robustness of SCAIR with respect to search hyperparameters, we evaluate performance under varying beam width $w$. We vary $w \in \{3,4,6\}$ while keeping the maximum depth fixed at $d=4$.

\begin{figure}[t]
    \centering
    \includegraphics[width=\columnwidth]{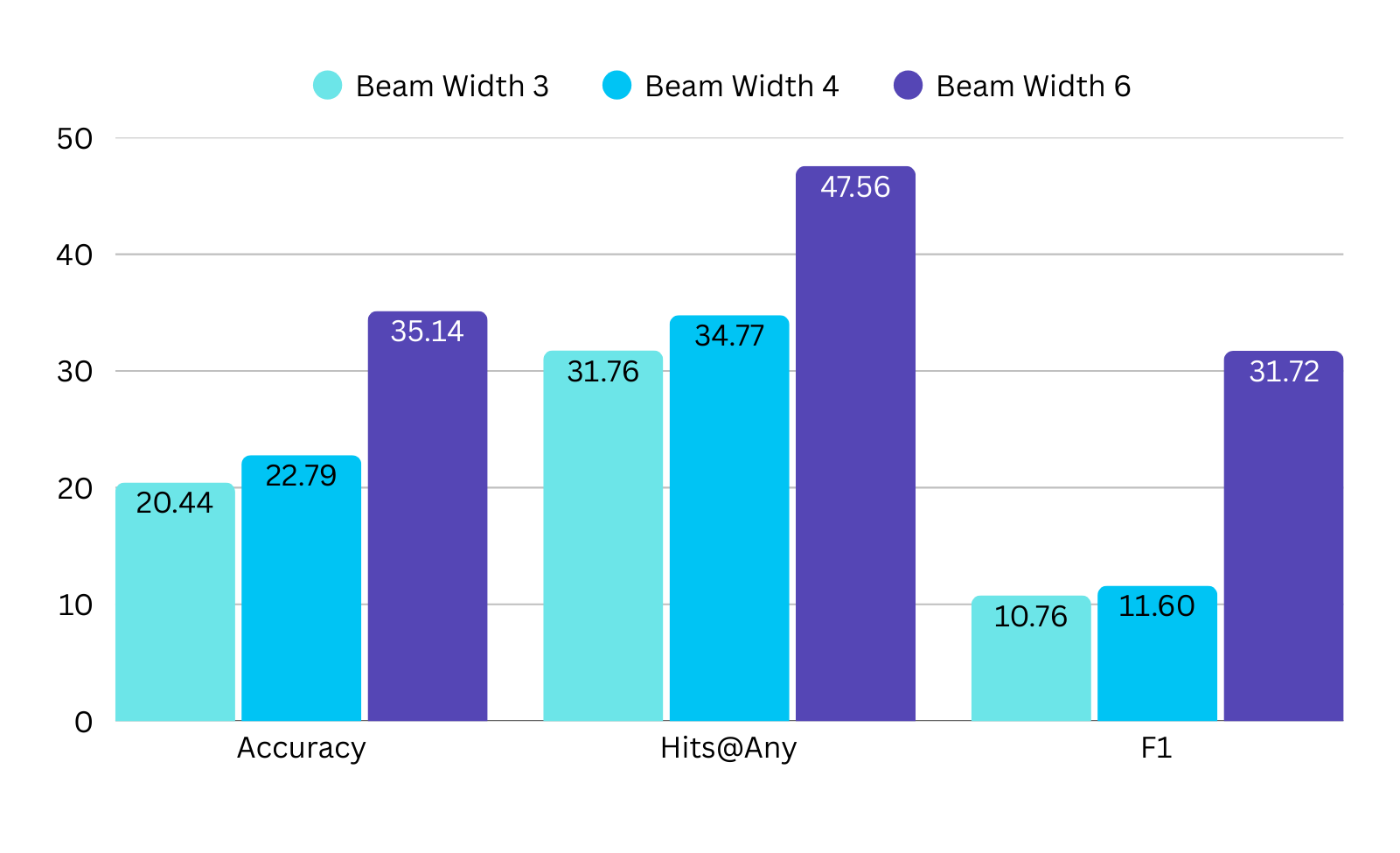}
    \caption{Performance of SCAIR under different beam widths.}
    \label{fig:beam_sweep}
\end{figure}

As shown in Figure~\ref{fig:beam_sweep}, increasing the beam width from 3 to 6 consistently improves Accuracy, Hits@Any, and F1. The improvement from $w=3$ to $w=4$ is moderate, while the gain from $w=4$ to $w=6$ is more pronounced, particularly for F1 and Hits@Any. This indicates that broader exploration improves coverage of relevant relations and entities in dense enterprise graphs.

However, the gains diminish as $w$ increases, suggesting that simply expanding the beam cannot indefinitely improve performance. Larger beams introduce more structurally irrelevant candidates, increasing computational cost without proportional accuracy gains. 

We therefore adopt $w=6$ as a balanced setting that provides strong performance while maintaining controlled exploration.

\subsection{Effect of Stronger Training-Based Backbones}
\label{app:qwen_experiment}

To investigate whether stronger language backbones can resolve the structural limitations observed in Plan\&Execute approaches, we replace the original LLaMA backbone in RoG with Qwen and retrain under identical settings.

In a controlled setting where gold-standard intermediate triples are provided, Qwen demonstrates stronger local reasoning ability than LLaMA, correctly answering complex compositional queries that LLaMA fails to resolve. This suggests that Qwen has improved conditional reasoning capacity when the relevant evidence is explicitly given.

\begin{figure}[t]
    \centering
    \includegraphics[width=\columnwidth]{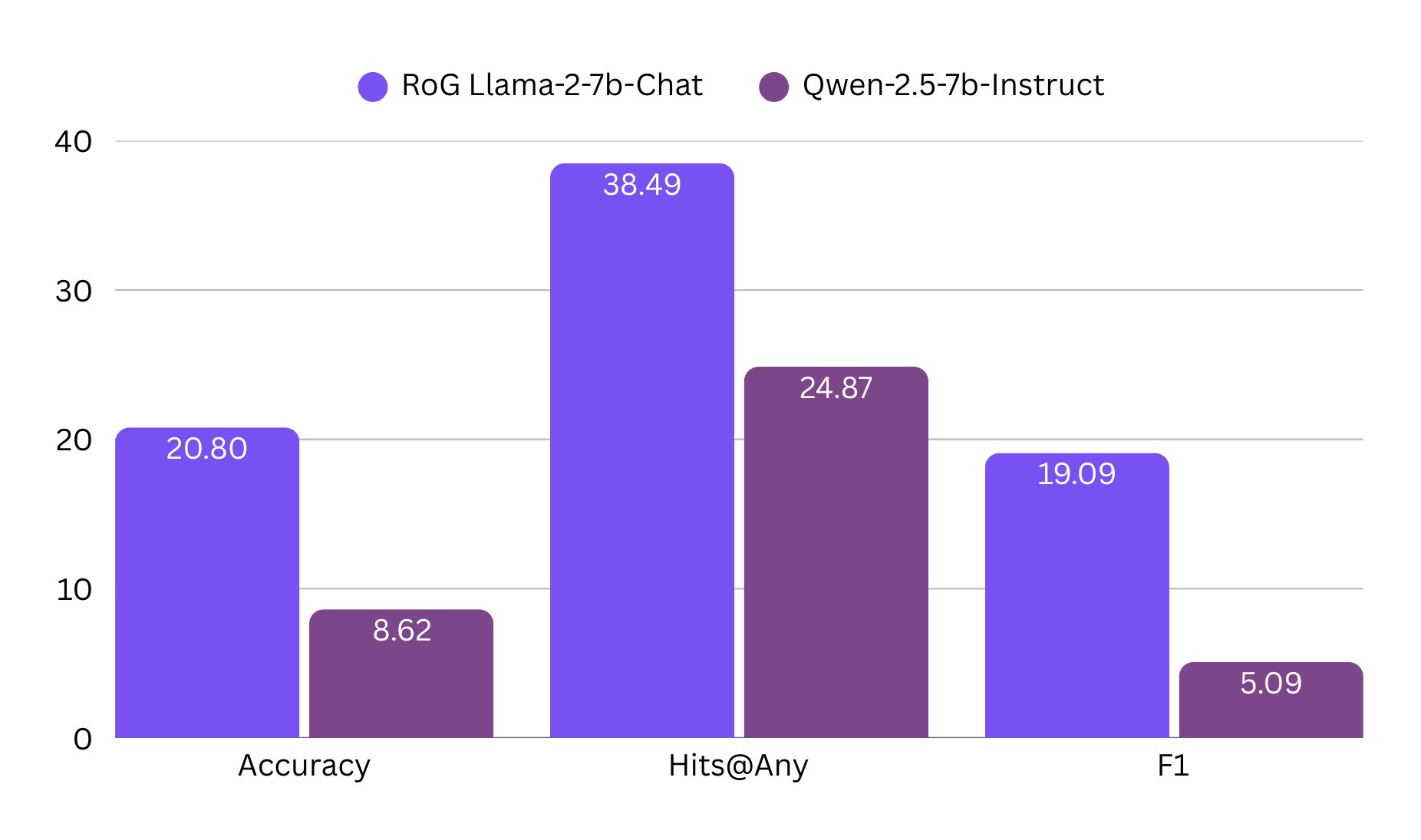}
    \caption{Comparison of RoG with LLaMA and Qwen backbones.}
    \label{fig:qwen_comparison}
\end{figure}

However, when evaluated in the full end-to-end KG-RAG pipeline, replacing LLaMA with Qwen does not yield meaningful improvements in overall benchmark performance (Figure~\ref{fig:qwen_comparison}). Performance on multi-hop and complex query types remains limited.

This discrepancy indicates that the primary bottleneck is not answer generation or conditional reasoning given correct evidence, but rather the structural search process that retrieves the evidence. Stronger LLM backbones improve reasoning over provided triples, but they do not mitigate uncontrolled traversal, schema-agnostic exploration, or search explosion in dense enterprise graphs.

These results reinforce our central claim: structural control over graph traversal is more critical than backbone strength for enterprise KG-RAG robustness.

\section{Extended Related Work}
\label{app:extended_related_work}

SCAIR addresses enterprise KG reasoning from the perspective of
schema-aware traversal control during agentic retrieval. For
completeness, we outline research threads that
complement this direction.

\paragraph{Embedding-based Retrieval.}
SCAIR retrieves evidence through entity grounding followed by
schema-constrained traversal. A complementary paradigm instead
embeds queries into a continuous space and retrieves answers via
similarity search over entity embeddings, turning multi-hop
reasoning into geometric composition rather than iterative
traversal \cite{DBLP:conf/iclr/RenHL20, DBLP:conf/iclr/ArakelyanDMC21, he2023can, he2024ecai, he2025dage}. 
The reliability of
these methods, however, is bounded by the quality of the
underlying KG embedding, which is itself uncertain \cite{zhu2025thesis}.
\citet{zhu-etal-2024-predictive} show that equally well-performing
KGE models systematically disagree on a large fraction of link
predictions, so a single point prediction cannot be treated as a
reliable summary of the retriever's output. To expose this
uncertainty explicitly, KGCP~\citep{zhu2025conformalized} and
CondKGCP~\citep{zhu-etal-2025-predicate} apply conformal prediction to
produce answer sets with marginal and predicate-conditional
coverage guarantees for KGE-based link prediction,
UnKGCP~\citep{zhu2025certainty} extends these guarantees to
uncertain KGEs through calibrated confidence intervals, and
BoxSEL~\citep{zhu2023towards, zhu2024approximating} provides approximate probabilistic
inference in Statistical $\mathcal{EL}$ with formal soundness
guarantees at the schema level. Until such statistical guarantees
are tightly integrated into embedding-based retrieval,
entity-grounded traversal as used in SCAIR remains a more
auditable choice for enterprise settings, where silently wrong
answers are costly.

\paragraph{Language Model Reasoning.}
An analogous tradeoff arises on the language-model side. One
could delegate a larger share of reasoning to the LLM, for
instance by letting it infer answers directly from loosely
retrieved context, but the LLM itself is a source of silent
unreliability. \citet{hesupposedly2025} show that entity
frequency in pretraining induces systematic asymmetries in LLM
factual reasoning, so that semantically equivalent facts are
handled inconsistently depending on the training distribution;
related instabilities also arise at the prompt level, where
minor wording changes can alter extracted outputs.
Complementary mitigations have been explored in prior work:
\citet{potyka2024robust} aggregate outputs across multiple
prompts through social choice theory, producing extractions that
are more stable than any single prompt, while
ArgRAG~\citep{zhu2025argrag} addresses the downstream reasoning
step by framing multi-document evidence combination as
quantitative bipolar argumentation, so that each generated claim
is backed by explicit supporting and attacking arguments.
Together with the retrieval-side analysis above, these results
support the central design choice of SCAIR: robust enterprise KG
reasoning requires jointly controlling the retrieval process and
the reasoning behavior of the underlying model, rather than
relying on either side alone.

\end{document}